\pdfoutput=1
\documentclass[11pt]{article}

\usepackage[]{acl}

\usepackage{times}
\usepackage{latexsym}
\usepackage{amsmath}
\usepackage{multicol}
\usepackage{graphicx}

\usepackage[T1]{fontenc}
\usepackage[utf8]{inputenc}

\usepackage{microtype}
\begin{document}
% If the title and author information does not fit in the area allocated, uncomment the following
%
%\setlength\titlebox{<dim>}
%
% and set <dim> to something 5cm or larger.

\title{Steno AI at SemEval-2023 Task 6: Rhetorical Role Labeling of Legal Documents using Transformers and Graph Neural Networks}

% Author information can be set in various styles:
% For several authors from the same institution:
% \author{Author 1 \and ... \and Author n \\
%         Address line \\ ... \\ Address line}
% if the names do not fit well on one line use
%         Author 1 \\ {\bf Author 2} \\ ... \\ {\bf Author n} \\
% For authors from different institutions:
% \author{Author 1 \\ Address line \\  ... \\ Address line
%         \And  ... \And
%         Author n \\ Address line \\ ... \\ Address line}
% To start a seperate ``row'' of authors use \AND, as in
% \author{Author 1 \\ Address line \\  ... \\ Address line
%         \AND
%         Author 2 \\ Address line \\ ... \\ Address line \And
%         Author 3 \\ Address line \\ ... \\ Address line}

% \author{Anshika Gupta \\ BITS Pilani \\ \texttt{f20200111@pilani.bits-pilani.ac.in}
%  \And Shaz Furniturewala \\ BITS Pilani \\ \texttt{f20200025@pilani.bits-pilani.ac.in}
%  \And Vijay Kunmari \\ BITS Pilani \\ \texttt{p20190065@pilani.bits-pilani.ac.in}
%  \And Yashvardhan Sharma \\ BITS Pilani \\ \texttt{yash@pilani.bits-pilani.ac.in}
% }

\author{\textbf{Anshika Gupta, Shaz Furniturewala, Vijay Kumari, Yashvardhan Sharma} \\
  BITS Pilani,Pilani,Rajasthan \\
  \texttt{(f20200111,f20200025)@pilani.bits-pilani.ac.in} \\
  \texttt{(p20190065,yash)@pilani.bits-pilani.ac.in} 
 } 

\maketitle

\begin{abstract}
A legal document is usually long and dense requiring human effort to parse it. It also contains significant amounts of jargon which make deriving insights from it using existing models a poor approach.
This paper presents the approaches undertaken to perform the task of rhetorical role labelling on Indian Court Judgements as part of SemEval Task 6:  understanding legal texts, shared subtask A \cite{legaleval-2023}. 
We experiment with graph based approaches like Graph Convolutional Networks and Label Propagation Algorithm, and transformer-based approaches including variants of BERT to improve accuracy scores on text classification of complex legal documents. \\
\end{abstract}

\section{Introduction}

Rhetorical Role Labelling for Legal Documents refers to the task of classifying sentences from court judgements into various categories depending on their semantic function in the document. This task is important as it not only has direct applications in the legal industry but also has the ability to aid several other tasks on legal documents such as summarization and legal search. This task is still in it's early stages, with huge scope for improvement over the current state-of-the-art.

To facilitate automatic interpretation of legal documents by dividing them into topic coherent components, a rhetorical role corpus was created for Task 6, sub-task A of The International Workshop on Semantic Evaluation \cite{legaleval-2023}. Several applications of legal AI, including judgment summarizing, judgment outcome prediction, precedent search, etc., depend on this classification.

\section{Related Works with Comparison}

The predominant technique used in Rhetorical Role Labeling over large datasets is based on the use of transformer-based models like LEGAL-BERT \cite{chalkidis-etal-2020-legal} and ERNIE 2.0 \cite{Sun_Wang_Li_Feng_Tian_Wu_Wang_2020}, augmented by various heuristics or neural network models. The accuracy of these approaches has remained low over the years. The results are summarized in Table 1.

The dataset \cite{aila-2021} used to implement the above approaches is relatively small, consisting only of a few hundred annotated documents and 7 sentence classes. 
\vspace{-5pt}
\begin{table}
\centering
\begin{tabular}{lc}
\hline
\textbf{Model} & \textbf{F1 score}\\
\hline
\verb|LEGAL-BERT| & {0.557} \\
\verb|LEGAL-BERT + Neural Net| & {0.517} \\
\verb|ERNIE 2.0| & {0.505} \\\hline
\end{tabular}
\begin{tabular}{lc}
\hline
\end{tabular}
\caption{Summary of related works on the task of rhetorical role labelling on legal text. \cite{aila-2021}}
\label{tab:accents}
\end{table}

\section{Dataset}
The dataset \cite{kalamkar-etal-2022-corpus} is made up of publicly available Indian Supreme Court Judgements. It consists of 244 train documents, 30 validation documents and 50 test documents making a total of 36023 sentences. \\
For every document, each sentence has been categorized into one of 13 semantic categories as follows:

\begin{enumerate}
    \item \textbf{PREAMBLE}: The initial sentences of a judgement mentioning the relevant parties
    \vspace{-7.5pt}
    \item \textbf{FAC}: Sentences that describe the events that led to the filing of the case
    \vspace{-7.5pt}
    \item \textbf{RLC}: Judgments given by the lower courts  based on which the present appeal was made to the present court
    \vspace{-7.5pt}
    \item \textbf{ISSUE}: Key points mentioned by the court upon which the verdict needs to be delivered
    \vspace{-7.5pt}
    \item \textbf{ARG\_PETITIONER}: Arguments made by the petitioner
    \vspace{-7.5pt}
    \item \textbf{ARG\_RESPONDENT}: Arguments made by the respondent
    \vspace{-7.5pt}
    \item \textbf{ANALYSIS}: Court discussion of the facts, and evidence of the case
    \vspace{-7.5pt}
    \item \textbf{STA}: Relevant statute cited
    \vspace{-7.5pt}
    \item \textbf{PRE\_RELIED}: Sentences where the precedent discussed is relied upon
    \vspace{-7.5pt}
    \item \textbf{PRE\_NOT\_RELIED}: Sentences where the precedent discussed is not relied upon
    \vspace{-7.5pt}
    \item \textbf{Ratio}: Sentences that denote the rationale/reasoning given by the Court for the final judgement
    \vspace{-7.5pt}
    \item \textbf{RPC}: Sentences that denote the final decision given by the Court for the case
    \vspace{-7.5pt}
    \item \textbf{None}: A sentence not belonging to any of the 12 categories
\end{enumerate}

\section{Proposed Techniques and Algorithms}

We try several different approaches for the task at hand. All our models use LEGAL-BERT as their base, and use various methods for further processing and refining of results.

The LEGAL-BERT family of models is a modified pretrained model based on the architecture of BERT \cite{devlin-etal-2019-bert}. The variant used in this paper is LEGAL-BERT-BASE, a model with 12 layers, 768 hidden units, and 12 attention heads. It has a total of 110M parameters and is pretrained for 40 epochs on a corpus of 12 GB worth of legal texts. 

This model was fine-tuned on the task dataset for 2 epochs with a learning rate of 1e-5 using the Adam optimizer and Cross entropy loss

\subsection{Direct Classification of CLS tokens}

First, we used the default classifier of LEGAL-BERT to find the first set of predictions, to establish a baseline for our further experiments. Our next step used the CLS tokens extracted from the final hidden layer of this trained model.

Similar to the methodology of Gaoa et al.\citeyearpar{gaoa2020legal} and Furniturewala et al.\citeyearpar{furniturewala2021legal} we utilised the CLS tokens from LEGAL-BERT for further classification models. This CLS token is a 768-dimensional semantic feature that represents BERT's understanding of the text input. It is a fixed embedding present as the first token in BERT's output to the classifier and contains all the useful extracted information present in the input text. 

We tried directly applying various multi-layer neural networks to the extracted CLS tokens. These two models served as a baseline to assess the efficacy of our methods.

\begin{figure}[h]
    \centering
    \includegraphics[scale=0.09]{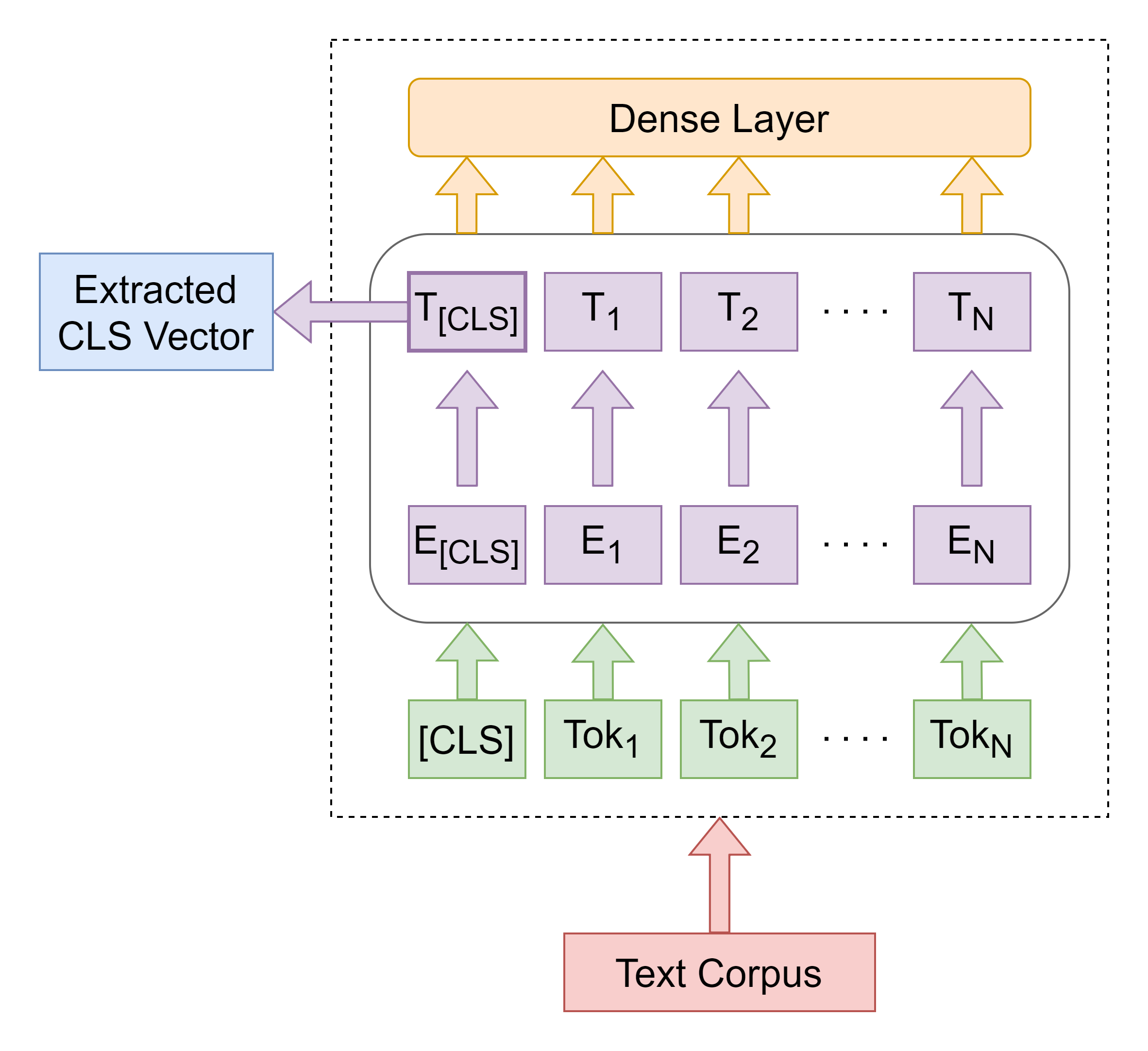}
    \caption{Extracting CLS Tokens (Furniturewala, 2021)}
    \label{fig:my_label}
\end{figure}

\subsection{Graph-Based Approaches}

We implemented classificaton systems based on graph architectures. We modeled the data into a graph using cosine similarity on the CLS tokens generated by LEGAL-BERT. An edge was created between two sentences if and only if their CLS tokens had cosine similarity greater than 0.5, with the cosine similarity acting as edge weight. The threshold was included to minimize the presence of noise-heavy edges in the graph.

\begin{equation} 
\cos ({\bf x},{\bf y}) = \frac{ \sum_{i=1}^{n}{{\bf x}_i{\bf y}_i} }{ \sqrt{\sum_{i=1}^{n}{({\bf x}_i)^2}} \sqrt{\sum_{i=1}^{n}{({\bf y}_i)^2}} } 
\end{equation}
\\ \\
The cosine similarity between two nodes, X and Y, is defined in equation (1), where x and y are the CLS tokens for nodes X and Y respectively, and n is the length of the CLS token, i.e. 768 in this case. The function for the final adjacency matrix is defined equation (2).

\vspace{-10pt}

\begin{equation}
  A_{XY} =
    \begin{cases}
      \cos ({\bf x},{\bf y}) & {if\hspace{2mm} \cos ({\bf x},{\bf y}) > 0.5}\\
      0 & {otherwise}
    \end{cases}       
\end{equation}

On this graph, we performed the label diffusion algorithm \cite{NIPS2003_87682805}, to establish a graph-based baseline for our system. Random walk label diffusion assigns labels to an unlabeled node using the average of it's neighbours, weighted by their distance from the node. 

% insert LPA formula here

\begin{equation}
    \centering{F^{t+1} = \alpha \cdot P \cdot F^t + (1 - \alpha) * Y}    
\end{equation}
\begin{equation}
    \centering{P = D^{-1/2}\cdot A\cdot D^{-1/2}}
\end{equation}
\begin{equation}
    \centering{F^* = (1 - \alpha) * (I - \alpha P)^{-1} \cdot Y}
\end{equation}

\begin{figure*}[h]
    \centering
    \includegraphics[scale=0.3]{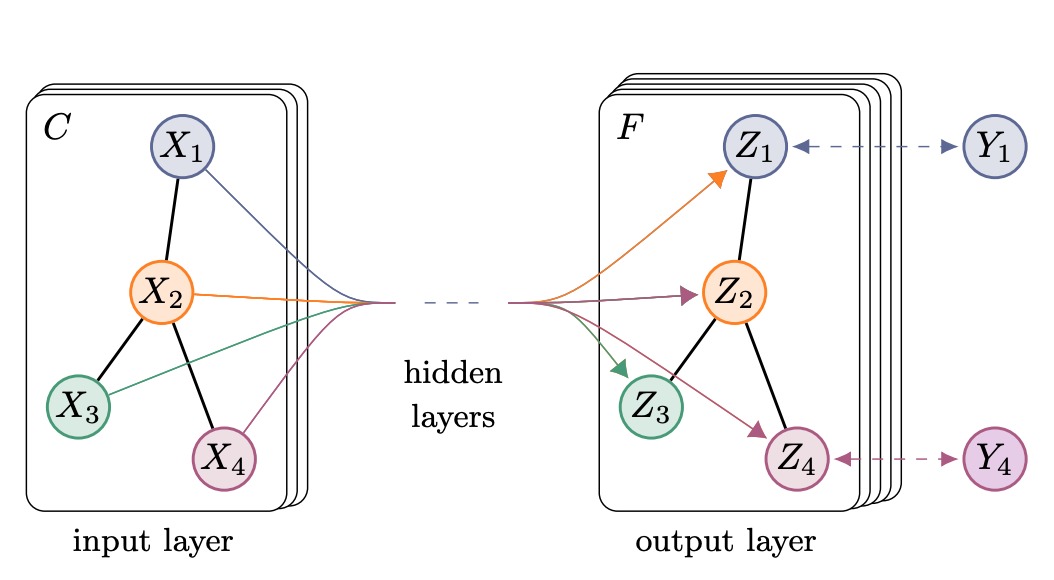}
    \caption{GCN Architecture \cite{gcn-kipf}}
    \label{fig:my_label}
\end{figure*}

To implement it, we combined the train and validation label array, one-hot encoded it and masked the validation labels. We then used equation (5) to generate predictions for each sentence. Here P is the normalised adjacency  matrix, Y is the array of one-hot encoded labels, $\alpha$ is a hyper-parameter, D is the degree matrix, and Z is the array of predicted labels.

The matrix P is obtained via equation (4), normalizing the adjacency matrix A using the square root inverse of the degree matrix D.
For our experimentation, we used $\alpha = 0.5$. 

 Furthermore, we used a two-layer Graph Convolution Network (GCN) \cite{gcn-kipf} to perform classifications on the data. Inspired by the methodology of BERTGCN \cite{lin-etal-2021-bertgcn}, we used the LEGAL-BERT embeddings of each sentence as the node representation for our graph, and then performed graph convolutions on it. 
 
 The GCN architecture uses trainable weights to identify the optimal weightage that each neighbour of each node should have on its label. The use of two layers allows us to incorporate the context of one-hop neighbours into the label of a particular node. 

\vspace{-10pt}

%insert GCN formula here
\begin{align}
% \begin{equation}
Z &= f(X, A)\\ &= softmax(\hat{A} \cdot ReLU(\hat{A}XW^{(0)})W^{(1)})
% \end{equation}
\end{align}

 We used equation (7) to predict the labels of the validation set. Here, Â represents the symmetrically normalized adjacency matrix, X is the feature vector which in this case is the LEGAL-BERT embeddings of the nodes, $W^i$ is the matrix of trainable weights in layer $i$. 
 
 The calculations required for this approach were extremely computationally expensive, so we were not able to train the model on the entire training set on a V100 server. We used half of the training documents for graph building and the prediction of labels. However, the LEGAL-BERT embeddings were generated by fine-tuning the model on all training documents.

\subsection{Context-Based LEGAL-BERT}

Our final approach was a Context-Based LEGAL-BERT. We cleaned each sentence by removing all stopwords (such as 'a', 'an', 'the') present using the NLTK library. Then we created a 5 sentence input corresponding to any given input by concatenating its two preceeding sentences and its two succeeding sentences in order. These 5 sentences were separated using LEGAL-BERT's separater token </s>. Sentences at the beginning or end of a document were padded using a string of <pad> tokens. 

These 5 sentence inputs were then tokenized using LEGAL-BERT's tokenizer and fed into the model using the baseline parameters. We used the default classifier to perform classification on these context-based inputs.

\section{Results}
We trained the models and tested them on the validation set. The accuracy scores have been reported in Table 2. 

We see that the performance of these models is significantly better than the previous attempts at this problem. The improvement of the results of previously studied models can be attributed to the increase in dataset size, along with other changes in the structure of the task. 

However, our Context-based LEGAL-BERT approach outperforms the other frameworks by a significant margin. This exhibits that the context of each sentence is critically important in determining its label, and that we are successful in incorporating the context of each sentence into its representation. 

We saw that graph-based approaches did not significantly improve performance compared to the current state-of-the-art models. However, it is important to note that we were unable to run the Graph Convolution Network using the entire train dataset due to compute constraints. 

Despite such constraints, there might be other reasons for the mediocre performance of graph-based models. One possible reason is that the representation of the sentences used for building the model was not able to capture information necessary to make better predictions. This also explains how the Context-based LEGAL-BERT performed so much better - it improved the quality of sentence representation, successfully capturing a wider range of features pertaining to the task at hand.

\begin{table}
\centering
\begin{tabular}{lc}
\hline
\textbf{Model} & \textbf{Accuracy}\\
\hline
\verb|LEGAL-BERT| & {65.56\%} \\
\verb|LEGAL-BERT + Classifier| & {67.15\%} \\
\verb|Graph Label Diffusion| & {66.34\%} \\
\verb|GCN| & {67.42\%} \\
\verb|Context-based LEGAL-BERT| & {71.02\%} \\\hline
\end{tabular}
\begin{tabular}{lc}
\hline
\end{tabular}
\caption{Summary of results obtained by the models on validation dataset}
\label{tab:acc}
\end{table}

\section{Conclusion and Future Work}

In this paper, we tried several different techniques to perform a sentence classification task on legal documents. Through our experiments, we show that incorporating context into the CLS tokens of sentences offers a significant improvement of 5.5 percentage points over LEGAL-BERT.

Moreover, through our experiments on graph-based models, we show that improving the CLS tokens results in a better classification, compared to the regular CLS tokens used in a variety of different ways. The Context-based LEGAL-BERT model was not only more accurate but also less resource intensive.

For future improvements on these models, we could try the Graph Convolutional Network approach on the complete dataset. We could also try the various methods of classification, such as a custom neural network or label diffusion, on the context-based CLS tokens. 

Moreover, we could further try to incorporate more sentences for context of each target sentence. This would require the use of a long-former model, since the total number of tokens passed into the model will increases. 

\bibliography{anthology, custom}
\bibliographystyle{acl_natbib}

\end{document}